\pdfoutput=1

\documentclass[11pt]{article}

\usepackage{EMNLP2023}

\usepackage{times}
\usepackage{latexsym}
\usepackage{booktabs}
\usepackage{bbm}

\usepackage[T1]{fontenc}

\usepackage[utf8]{inputenc}

\usepackage{microtype}

\usepackage{inconsolata}
\usepackage{multirow}
\usepackage{color}
\usepackage{graphicx}
\usepackage{amsmath}
\usepackage{subfigure}

\title{DAIL: Data Augmentation for In-Context Learning via Self-Paraphrase}

\author{Dawei Li, Yaxuan Li, Dheeraj Mekala, Shuyao Li, \\
\textbf{Yulin wang, Xueqi Wang, William Hogan, Jingbo Shang} \\
        University of California, San Diego \\
        \texttt{dal034, yal105, dmekala, shl118, yuw033} \\ 
        \texttt{xuw030, whogan, jshang@ucsd.edu}}

\begin{document}
\maketitle
\begin{abstract}
In-Context Learning (ICL) combined with pre-trained large language models has
achieved promising results on various NLP tasks.
However, ICL requires high-quality annotated demonstrations which might not be available in real-world scenarios.
To overcome this limitation, we propose \textbf{D}ata \textbf{A}ugmentation for \textbf{I}n-Context \textbf{L}earning (\textbf{DAIL}).
DAIL leverages the intuition that large language models are more familiar with the content generated by themselves.
It first utilizes the language model to generate paraphrases of the test sample and employs majority voting to determine the final result based on individual predictions.
Our extensive empirical evaluation shows that DAIL outperforms the standard ICL method and other ensemble-based methods in the low-resource scenario.
Additionally, we explore the use of voting consistency as a confidence score of the model when the logits of predictions are inaccessible. 
We believe our work will stimulate further research on ICL in low-resource settings.
\end{abstract}

\label{method}

\begin{figure*}[!t]
    \centering
    \includegraphics[width=15cm]{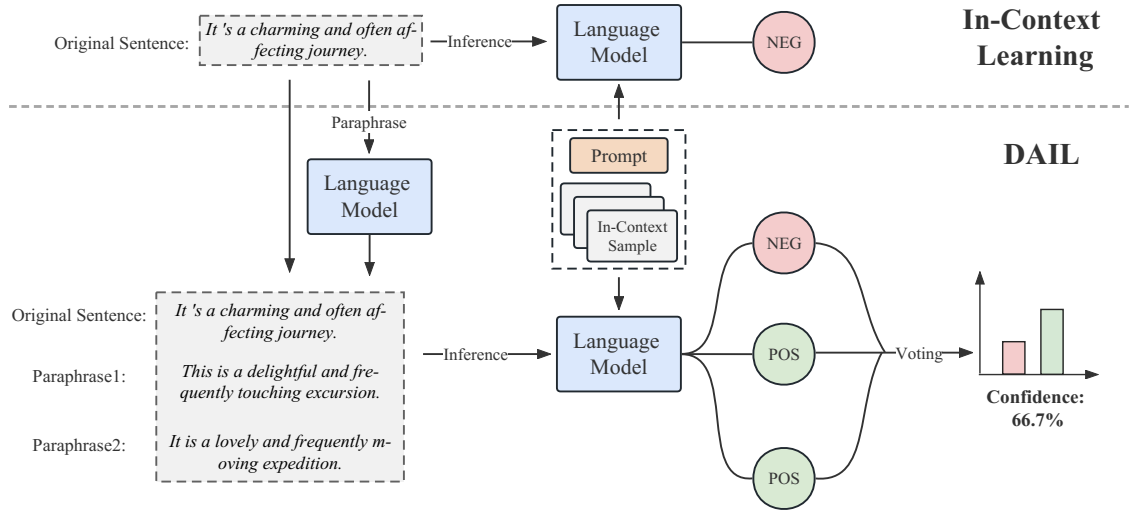}

    \caption{An overview pipeline of our method, DAIL. The part above the dotted line represents the standard In-Context Learning process, and the lower part represents our method. DAIL augments the test sample by generating multiple paraphrases, obtains individual predictions, and ensembles them to return the final answer.}
    \label{pipeline}
\end{figure*}

\section{Introduction}
Recently, the rapid development of large language models (LLMs)~\cite{devlin2018bert,radford2019language} and their striking skill and knowledge have sparked significant interest in In-Context Learning (ICL)~\cite{brown2020language}.
Different from other training-based paradigms, under ICL, there is no parameter adjustment to LLMs.
Instead, the model is given an instruction, which usually consists of a task description (or prompt), several in-context samples (or demonstration), and the test case that needs to be inferred.
This tuning-free approach has gained prominence in various research domains within Natural Language Processing (NLP) due to its impressive performance in numerous downstream tasks~\cite{bonifacio2022inpars,gao2023chat,Gao2023HumanlikeSE}.



ICL requires demonstrations as part of the context.
Many works focus on improving ICL by employing diverse methods to select suitable demonstrations, such as similarity-based retrieval~\cite{liu2021makes,rubin2021learning}, score function learning~\cite{li2023finding}, permutation search~\cite{lu2021fantastically,wu2022self}, and etc. 
However, in numerous real-world scenarios, the availability of annotated samples is severely constrained~\cite{mekala2020contextualized, su2022selective} for such demonstration selection.
Moreover, in the fine-grained classification with a large target-label space, the context length of LLMs could not accommodate more than one demonstration per label.

To address these problems, we leverage the insight of using the ensemble method and improve ICL for low-resource scenarios.
We propose \textbf{D}ata \textbf{A}ugmentation for \textbf{I}n-Context \textbf{L}earning (\textbf{DAIL}) that uses the large language model paraphrase the test sample multiple times.
Subsequently, the paraphrased samples, along with the original test sample, are separately presented to the large language model for inference.
Finally, a majority voting mechanism is employed to integrate all the single results and get the final label.
In the experiments, we show the effectiveness and generalization of our proposed method by evaluating on coarse-grained and fine-grained classification tasks.
Additionally, we demonstrate the potential to utilize voting consistency to serve as the confidence score of the model output when the logits are inaccessible, such as for closed LLMs (e.g. ChatGPT).

Note that \citet{wei2022chain, wang2022self} employ ensembling methods for reasoning.
These studies employ various sampling methods to obtain multiple reasoning paths of the model and generate answers by ensembling them.
However, in Section~\ref{Ensemble-based Methods}, we demonstrate that their effectiveness is heavily reliant on providing well-prepared reasoning samples to the model's input.

\section{DAIL: Data Augmentation for In-Context Learning}

The intuition behind our method is that
LLM is more familiar with what it generates i.e.
it would likely have better performance while inferencing with the text generated by itself. 
Here, we refer to this as \emph{self-paraphrase}.
We validate this empirically in Section~\ref{sec:results} and observe it to be true.

Figure~\ref{pipeline} shows the overview pipeline of our DAIL. 
First, in line with the self-paraphrase assumption, we augment the data by paraphrasing the original text of the test sample using the LLM itself and getting multiple candidates for the next step.
Then, each candidate is concatenated with a prompt corresponding to each task, along with $m$ randomly selected demonstrations from the training set.
To mimic the low-resource scenario, we consider no more than $1$ demonstration per label.
The final label is generated by considering the candidates from all the paraphrased texts in addition to the original text.  
This is achieved through a majority voting approach, where the final label is determined based on the majority consensus among the candidates. 
Finally, for LLMs whose logits are inaccessible, we consider the voting consistency as its confidence score.

Mathematically, DAIL obtains $n$ samples by paraphrasing the test sample $s$:
\begin{equation}
    S_{p} = {\rm LLM}_{para}(s, n)
\end{equation}
Then, DAIL uses the $n$ paraphrased samples together with the original sample to perform inference independently and get $n+1$ labels in total:
\begin{equation}
    a_i = {\rm LLM}_{inf}(p;d;s_i), \quad s_i \in S_{p} \cup \{s\}
\end{equation}
Here $p$ and $d$ represent the prompt and demonstrations respectively.
Assume the total label space is $L$, then the final result through majority voting is:
\begin{equation}
    a_{vote} = \underset{l \in L}{\mathrm{argmax}}\sum_{i=1}^{n + 1} \mathbbm{1} (a_i = l)
\end{equation}
Finally, we consider voting consistency as the confidence score using:
\begin{equation}
    c = \frac{\sum_{i=1}^{n + 1} \mathbbm{1} (a_i = a_{vote})}{n + 1}
\end{equation}

\begin{table*}[]\small
\centering
\begin{tabular}{cccccccccc}
\hline
Model                      & Method              & SST2 & SST5 & CR   & Emotion & TREC & AG News & Emp Dialogue & Yahoo\\ \hline
\multirow{5}{*}{ChatGPT}   & Standard ICL        & 93.6 & 54.4 & 88.0 & 52.2    & 77.1 & 84.6  &  43.1 &  64.3 \\ \cline{2-10} 
                           & DAIL-1              & 93.8 & 52.2 & 89.4 & 52.4    & 79.6  & 84.7 & 43.0  & 69.6 \\ \cline{2-10} 
                           & DAIL-2              & 93.6 & 54.2 & 89.6 & 54.1    & 82.5 & 87.4 & 45.5  & 70.6 \\ \cline{2-10} 
                           & DAIL-4               & 93.6 & \bf{55.4} & 89.9 & \bf{54.6}    & \bf{84.6} & \bf{87.6} & \bf{45.7}  & 70.3 \\ \cline{2-10} 
                           & DAIL-4 w/o SP    & \bf{94.7} & 53.8 & \bf{90.3} & 51.5    & 82.1 & 85.2  & 42.2  & \bf{71.2} \\ \hline
\multirow{5}{*}{PaLM2-540B} & Standard ICL        & 93.2 & 48.7 & 91.7 & 50.5    & 70.1  & 86.2  & 51.4  & 67.5 \\ \cline{2-10} 
                           & DAIL-1              & 92.5 & 52.3 & 91.4 & 50.5    & 69.6 & 84.1  & 49.8  & 67.9  \\ \cline{2-10} 
                           & DAIL-2              & 93.4 & 51.8 & 91.1 & 52.2    & 73.9 & 87.2  & 51.9  & 68.4  \\ \cline{2-10} 
                           & DAIL-4               & \bf{93.4} & \bf{54.8} & \bf{92.3} & \bf{53.0}    & \bf{76.4} & \bf{88.6}   & \bf{53.0}  & \bf{68.5} \\ \cline{2-10} 
                           & DAIL-4 w/o SP & 93.2 & 52.9 & 92.3 & 52.1    & 73.3 & 85.5  & \textbf{53.6}  & 67.7  \\ \hline

\end{tabular}
\caption{Comparison of DAIL and Standard ICL. We report the accuracy score for each dataset. The best results are in bold. Our DAIL-4 achieves the overall best results.}
\label{table:mainResults}
\end{table*}

\section{Experiments}

\subsection{Experimental Settings}

We use several classification tasks in our experiments, including SST2, SST5~\cite{socher2013recursive}, CR~\cite{hu2004mining}, Emotion~\cite{saravia2018carer}, TREC~\cite{voorhees2000building} and AGNews~\cite{zhang2015character}.
We also use two fine-grained classification datasets, Empathetic Dialogues~\cite{rashkin2018towards} and Yahoo Answer Topics~\cite{sileo2023tasksource} to evaluate DAIL's performance with large label spaces.
Table~\ref{dataset stat} shows the detailed statistics of the aforementioned datasets.
For baseline models, we choose ChatGPT\footnote{https://platform.openai.com/docs/models/gpt-3-5} and PaLM2-540B~\cite{anil2023palm}.
We use the official APIs of each model to conduct evaluations.
For ChatGPT, we use a stable version \texttt{GPT-3.5-turbo-0301} for our experiments.
Due to the lack of a stable version of the PaLM2-540B model, we repeat the experiments three times and report the average results.
To simulate a low-resource scenario, we establish the number of demonstrations for each label category in each dataset to be 1.
We report the accuracy score for each dataset.

We compare our method with standard ICL.
For our method, we compare several numbers of paraphrases per sample ranging from 1 to 4 and DAIL-$i$ indicates $i$ paraphrases per sample.
For DAIL-1, we simply replace the original test sample with a paraphrase generated by the LLM and perform inference with it.
For DAIL-2 and DAIL-4, we follow the method we illustrate in Section~\ref{method} to generate and ensemble paraphrases.
More details about our paraphrase and prompt format can be found in Appendix~\ref{Prompt for DAIL}

\begin{table}[]\small
\centering
\begin{tabular}{ccc}
\hline
Task                                      & Dataset        & Testset Size \\ \hline
\multirow{3}{*}{Sentiment Classification} & SST2           & 872          \\ \cline{2-3} 
                                          & SST5           & 2210         \\ \cline{2-3} 
                                          & CR             & 376          \\ \hline
\multirow{3}{*}{Topic Classification}     & AGNews         & 7600         \\ \cline{2-3} 
                                          & TREC           & 500          \\
\cline{2-3} 
                                          & Yahoo          & 2000          \\
                                          \hline
\multirow{2}{*}{Emotion Classification}   & Emotion        & 2000         \\ \cline{2-3} 
                                          & Emp Dialogue   & 10900          \\ \hline
\end{tabular}
\caption{Detailed statistics of the datasets we use in the experiment.}
\label{dataset stat}
\end{table}

\subsection{Main Results}
\label{sec:results}
Table~\ref{table:mainResults} shows the comparison between DAIL and standard ICL method.
To further validate our assumption underlying the self-paraphrase mechanism, we also conduct an ablation study on DAIL-4 by using paraphrases interchangeably, denoted by DAIL-4 w/o SP.
We use the paraphrase generated by PaLM for ChatGPT and vice versa.

Here DAIL-1 doesn't exhibit a significant improvement in all datasets, suggesting the necessity of adopting an ensemble-based approach in DAIL.
When the number of paraphrase samples used in DAIL increases, the performance consistently improves.
DAIL-4 attains the best overall results across various datasets for both models.
For datasets such as Yahoo that have $10$ target classes, the performance improvement is up to $6$ points in the case of ChatGPT.
It is also notable that the performance in both models drops when we replace the self-paraphrase mechanism in DAIL-4.
This confirms the assumption we put forth that large language models are more accustomed to the content they generate.

\begin{table}[h!]\small
\centering
\begin{tabular}{cccc}
\hline
                                      & SST5                                                & TREC                                                & \multicolumn{1}{l}{Emotion}                        \\ \hline
Standard ICL                          & 54.4 &  77.1 & 52.2 \\ \hline
\begin{tabular}[c]{@{}c@{}}Self-Consistency\\ \cite{wang2022self}\end{tabular}                      & 54.8                        & 78.2                        & \bf{55.6}                        \\ \hline
\begin{tabular}[c]{@{}c@{}}Prompt-Ensemble\\ \cite{gao2020making}\end{tabular} & 54.2                        & 81.9                        & 55.5                        \\ \hline
DAIL                                  & \bf{55.4}                        & \bf{84.6}                        & \bf{55.6}                        \\ \hline
\end{tabular}
\caption{Comparison of DAIL and other ensemble-based methods. The best results are in bold. We observe DAIL out-performing other ensemble-based methods for classification.}
\label{table:comparisonToOtherEnsembleBasedMethods}
\end{table}

\subsection{Comparison to Other Ensemble-based Methods}
\label{Ensemble-based Methods}

\begin{table*}[]\small
\centering
\begin{tabular}{c|c|c|c}
\hline
                      & Text                                                                            & Prediction & \multicolumn{1}{l}{Ground Truth} \\ \hline
Org                   & A \textcolor[RGB]{76,175,80}{well acted} and \textcolor[RGB]{76,175,80}{well intentioned} \textcolor[RGB]{244,67,54}{snoozer}.                                      & Neutral    & \multirow{5}{*}{Negative}         \\ \cline{1-3}
\multirow{4}{*}{Para} & Although \textcolor[RGB]{76,175,80}{well-performed} and \textcolor[RGB]{76,175,80}{well-meaning}, it can be quite \textcolor[RGB]{244,67,54}{dull}.                 & Negative   &                                   \\ \cline{2-3}
                      & It's a \textcolor[RGB]{244,67,54}{yawn-inducing} film, but the acting and themes are \textcolor[RGB]{76,175,80}{commendable}.           & Neutral    &                                   \\ \cline{2-3}
                      & The movie is a \textcolor[RGB]{244,67,54}{bore}, despite the \textcolor[RGB]{76,175,80}{admirable} acting and \textcolor[RGB]{76,175,80}{good} intentions.          & Negative   &                                   \\ \cline{2-3}
                      & While \textcolor[RGB]{76,175,80}{well-acted} and with \textcolor[RGB]{76,175,80}{good intentions}, the movie is a bit of a \textcolor[RGB]{244,67,54}{snooze fest}. & Negative   &                                   \\ \hline
\end{tabular}
\caption{An example in SST5 dataset. The text in \textcolor[RGB]{76,175,80}{green} represent positive content and the text in \textcolor[RGB]{244,67,54}{red} represent negative content}
\label{case study}
\end{table*}

We conduct a comparison between our method and two other ensemble-based approaches, namely Self-Consistency~\cite{wang2022self} and Prompt-Ensemble~\cite{gao2020making}. 


Self-Consistency is a method that incorporates sampling different reasoning paths during decoding.
The various sampled results are subsequently used for voting to determine the final result. 
Prompt-Ensemble~\cite{gao2020making}, on the other hand, employs different prompts for multiple inferences and then uses the outputs of the model to determine the final prediction through voting. 
In our experiments, we manually write prompts for each dataset.
More details about it can be found in Appendix~\ref{Prompt for Prompt-Ensemble}.

Table~\ref{table:comparisonToOtherEnsembleBasedMethods} displays the comparison results.
While the three methods demonstrate similar performance on the Emotion dataset, DAIL exhibits a significant advantage over the other two methods in terms of SST5 and TREC.
Additionally, we observe the performance of Self-Consistency in the classification datasets we use is not as strong as reported in the original paper for several reasoning datasets.
One possible explanation for this discrepancy is that the classification datasets we employ lack reasoning samples that are provided to the model.
In contrast, DAIL does not rely on any external sources, making it a more comprehensive approach.




\begin{figure}[t]
    \subfigure[Binary classification]{
        \includegraphics[width=0.47\linewidth]{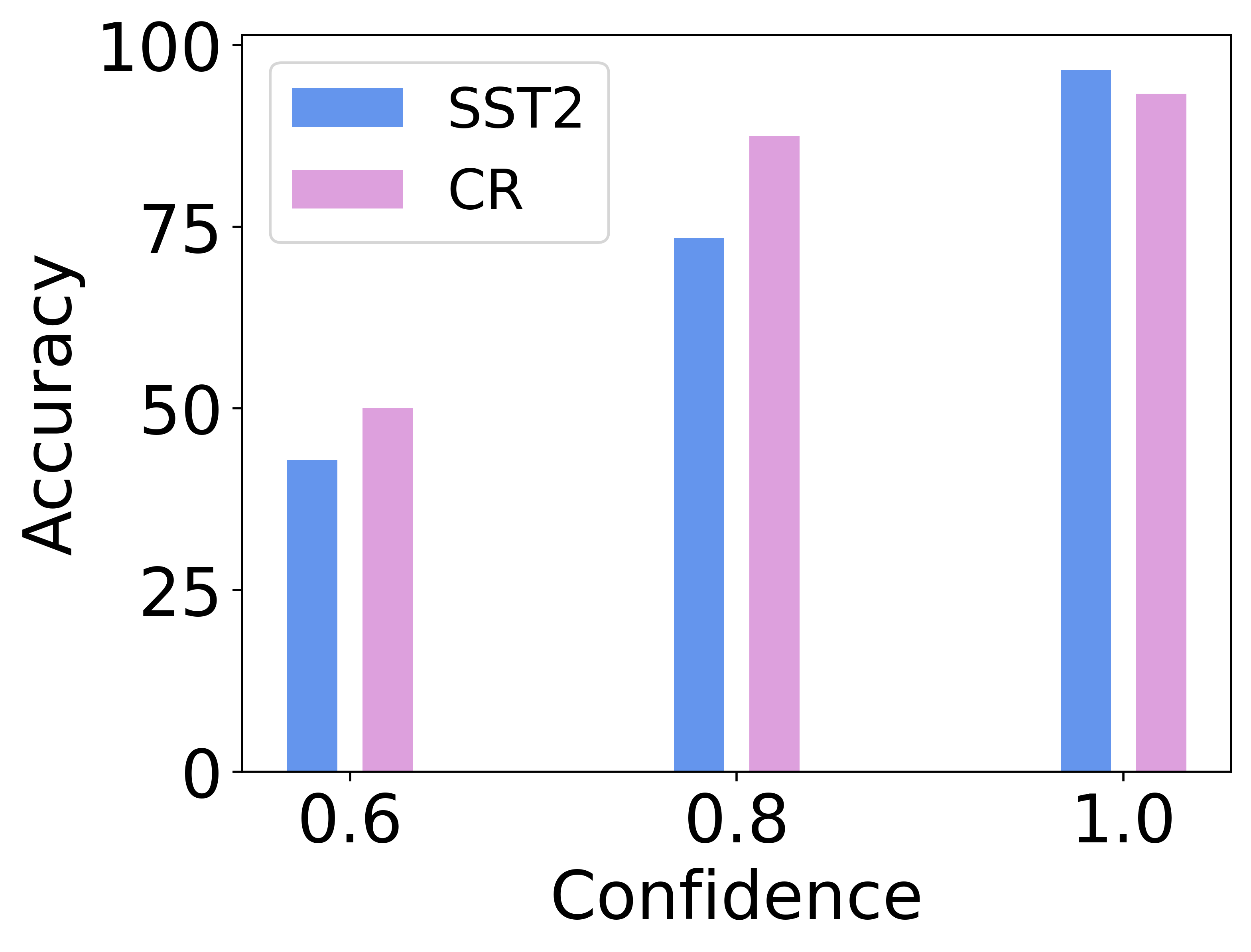}
    }
    \subfigure[Multi-class classification]{
        \includegraphics[width=0.47\linewidth]{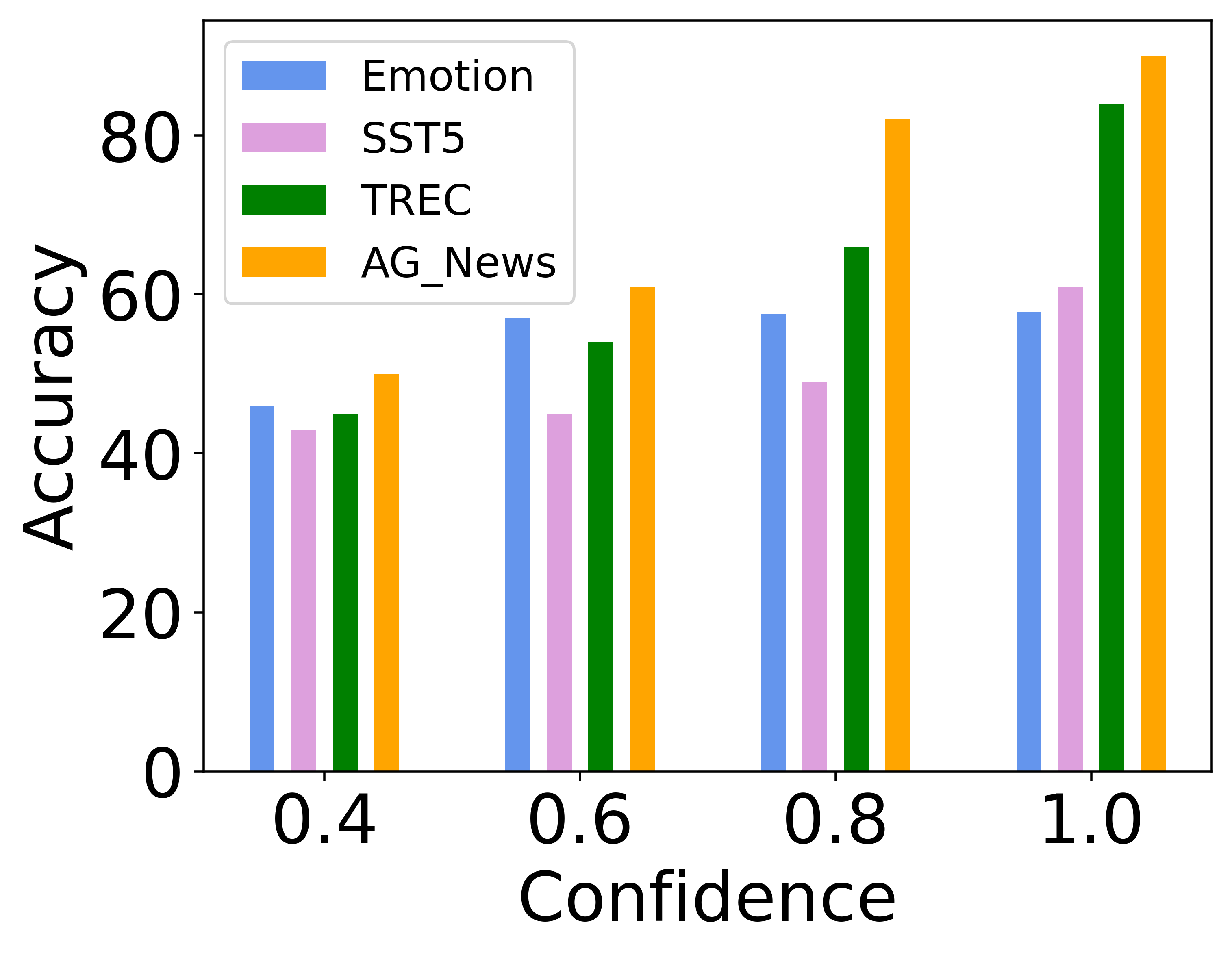}
    }
    \vspace{-3mm}
    \caption{Model Accuracy vs Voting consistency score using ChatGPT. We observe a positive correlation between voting consistency score and accuracy, thus making it a reliable confidence metric.} 
    \label{Voting cosistency VS model accuracy on ChatGPT}
    \vspace{-5mm}
\end{figure}

\subsection{Voting Consistency as Confidence Score}

In this section, we assess the reliability of employing voting consistency as the confidence score for the model. 
We plot the accuracy of ChatGPT vs several voting consistency thresholds illustrating their relationship in Figure~\ref{Voting cosistency VS model accuracy on ChatGPT}.
For the binary classification datasets (SST2, CR), we utilize confidence values of 0.6, 0.8, and 1.0. 
For the multiple classification datasets (SST5, Emotion, TREC, AGNews), we use confidence values of 0.4, 0.6, 0.8, and 1.0.
We put the results of PaLM in Appendix~\ref{Voting Consistency as Confidence Score: PaLM}.




Although the confidence score doesn't completely correspond to the accuracy, 
we observe a significant positive correlation with each other.
By considering the confidence level, we can roughly assess the reliability of the model's output.
We believe this could serve as a confidence score for addressing the problems in LLMs such as hallucination~\cite{ji2023survey} and confidence calibration~\cite{wang2020inference, mekala-etal-2022-lops,  mekala2023selfood}.

\subsection{Case Study}
To explain our method, we provide a case study from ChatGPT's results on SST5.
Consider a negative sentiment sentence: ``\textit{A well acted and well intentioned snoozer.}''.
Here both positive expression (``\textit{well acted}'' and ``\textit{well intentioned}'') and negative expressions (``\textit{snoozer}'') are present, making it a challenging one for predicting sentiment.

For standard ICL, the model fails to capture this subtle sentiment and gives the wrong label 'Neutral'.
The paraphrases generated by DAIL are mentioned in Table~\ref{case study}.
For DAIL, one of the paraphrases generated by ChatGPT is ``\textit{Although well-performed and well-meaning, it can be quite dull}''.
It emphasizes the switch in relationship with the word ``\textit{although}'' and gets the correct label 'Negative'.
By paraphrasing several times and using majority voting, DAIL predicts it correctly.




\section{Conclusion}

In this work, we introduce a data augmentation technique to ICL and propose DAIL.
It involves augmenting the test sample by generating multiple paraphrases and ensembling the individual results to obtain the final prediction.
In our experiments, we compare DAIL with standard ICL and other ensemble-based methods. 
The evaluation results demonstrate the effectiveness of DAIL in the low-resource ICL scenario.
Furthermore, we investigate the feasibility of utilizing voting consistency as a confidence estimation method and discover a positive correlation between voting consistency and model accuracy.

\section{Limitations}
Similar to other ensemble-based methods, DAIL requires multiple inferences to obtain individual labels for each paraphrased sample. 
This introduces additional costs compared to standard ICL. 
Additionally, our self-paraphrase mechanism relies on the same language model to generate paraphrases for the test sample. 
Consequently, DAIL may not be suitable for smaller language models that lack the capability to produce high-quality paraphrases.

\bibliography{anthology,custom}
\bibliographystyle{acl_natbib}

\appendix
\onecolumn

\section{Prompt for DAIL}
\label{Prompt for DAIL}

We present the prompt we use to generate paraphrases and do inference with LLMs in Table~\ref{Paraphrase Prompt} and Table~\ref{Inference Prompt}. We utilize <Label Space> to represent all labels in certain datasets, <Demonstrations> to represent the demonstrations used, and <Test Sample> to represent the sample being inferred.

\begin{table}[h!]\tiny
\centering
\begin{tabular}{c|c|l}
\hline
Task                                      & Dataset             & \multicolumn{1}{c}{Paraphrase Prompt}                                                                                          \\ \hline
\multirow{3}{*}{Sentiment Classification} & SST2                & \multirow{3}{*}{Please paraphrase this sentence \textless{}Para-Num\textgreater times without changing the original sentiment:} \\ \cline{2-2}
                                          & SST5                &                                                                                                                                 \\ \cline{2-2}
                                          & CR                  &                                                                                                                                 \\ \hline
\multirow{2}{*}{Emotion Classification}   & Emotion             & \multirow{2}{*}{Please paraphrase this sentence \textless{}Para-Num\textgreater times without changing the original emotion:}   \\ \cline{2-2}
                                          & Empathetic Dialogue &                                                                                                                                 \\ \hline
\multirow{2}{*}{Question Classification}  & TREC                & \multirow{2}{*}{Please paraphrase this question \textless{}Para-Num\textgreater times without changing the original semantic:}  \\ \cline{2-2}
                                          & Yahoo               &                                                                                                                                 \\ \hline
News Classification                       & AG News             & Please paraphrase this news \textless{}Para-Num\textgreater times without changing the original semantic:                       \\ \hline
\end{tabular}
\caption{Prompt we use to generate paraphrases of the test samples for each dataset.}
\label{Paraphrase Prompt}
\end{table}

\begin{table}[h!]\tiny
\centering
\begin{tabular}{c|c|l}
\hline
Task                                      & Dataset             & \multicolumn{1}{c}{Inference Prompt}                                                                                                                                                                                                  \\ \hline
\multirow{3}{*}{Sentiment Classification} & SST2                & \multirow{3}{*}{\begin{tabular}[c]{@{}l@{}}Label the sentiment class of the sentence,\\ please choose from \textless{}Label Space\textgreater \textless{}Demonstrations\textgreater \textless{}Test Sample\textgreater{}\end{tabular}} \\ \cline{2-2}
                                          & SST5                &                                                                                                                                                                                                                                        \\ \cline{2-2}
                                          & CR                  &                                                                                                                                                                                                                                        \\ \hline
\multirow{2}{*}{Emotion Classification}   & Emotion             & \multirow{2}{*}{\begin{tabular}[c]{@{}l@{}}Label the emotion class of the sentence,\\ please choose from \textless{}Label Space\textgreater \textless{}Demonstrations\textgreater \textless{}Test Sample\textgreater{}\end{tabular}}   \\ \cline{2-2}
                                          & Empathetic Dialogue &                                                                                                                                                                                                                                        \\ \hline
\multirow{2}{*}{Question Classification}  & TREC                & \multirow{2}{*}{\begin{tabular}[c]{@{}l@{}}Label the category of the question,\\ please choose from \textless{}Label Space\textgreater \textless{}Demonstrations\textgreater \textless{}Test Sample\textgreater{}\end{tabular}}        \\ \cline{2-2}
                                          & Yahoo               &                                                                                                                                                                                                                                        \\ \hline
News Classification                       & AG News             & \begin{tabular}[c]{@{}l@{}}Label the category of the news,\\ please choose from \textless{}Label Space\textgreater \textless{}Demonstrations\textgreater \textless{}Test Sample\textgreater{}\end{tabular}                             \\ \hline
\end{tabular}
\caption{Prompt we use to do inference for each dataset.}
\label{Inference Prompt}
\end{table}

\section{Voting Consistency as Confidence Score: PaLM}
\label{Voting Consistency as Confidence Score: PaLM}

\begin{figure}[h]
    \subfigure[Binary classification]{
        \includegraphics[width=0.47\linewidth]{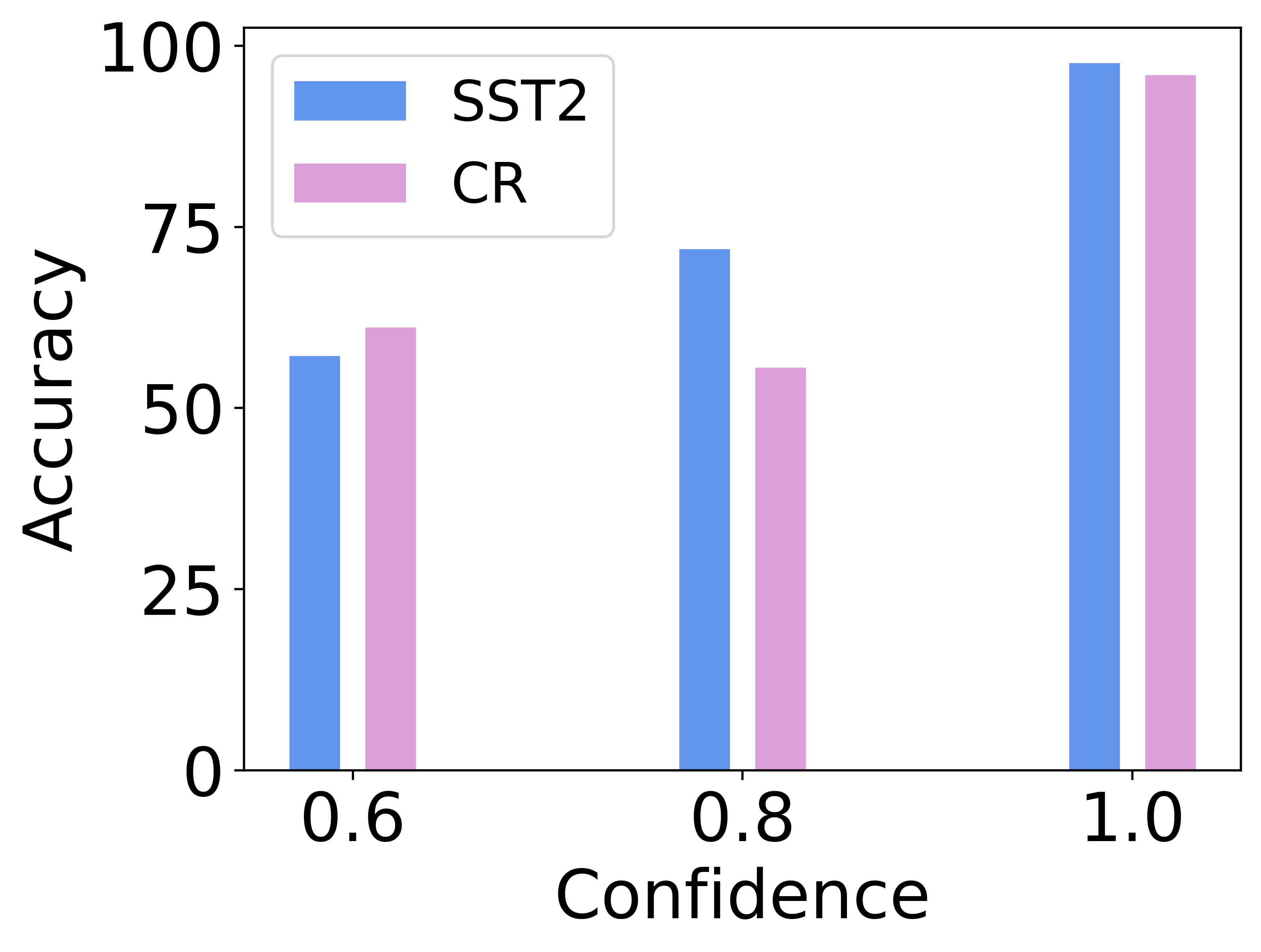}
    }
    \subfigure[Multi-class classification]{
        \includegraphics[width=0.47\linewidth]{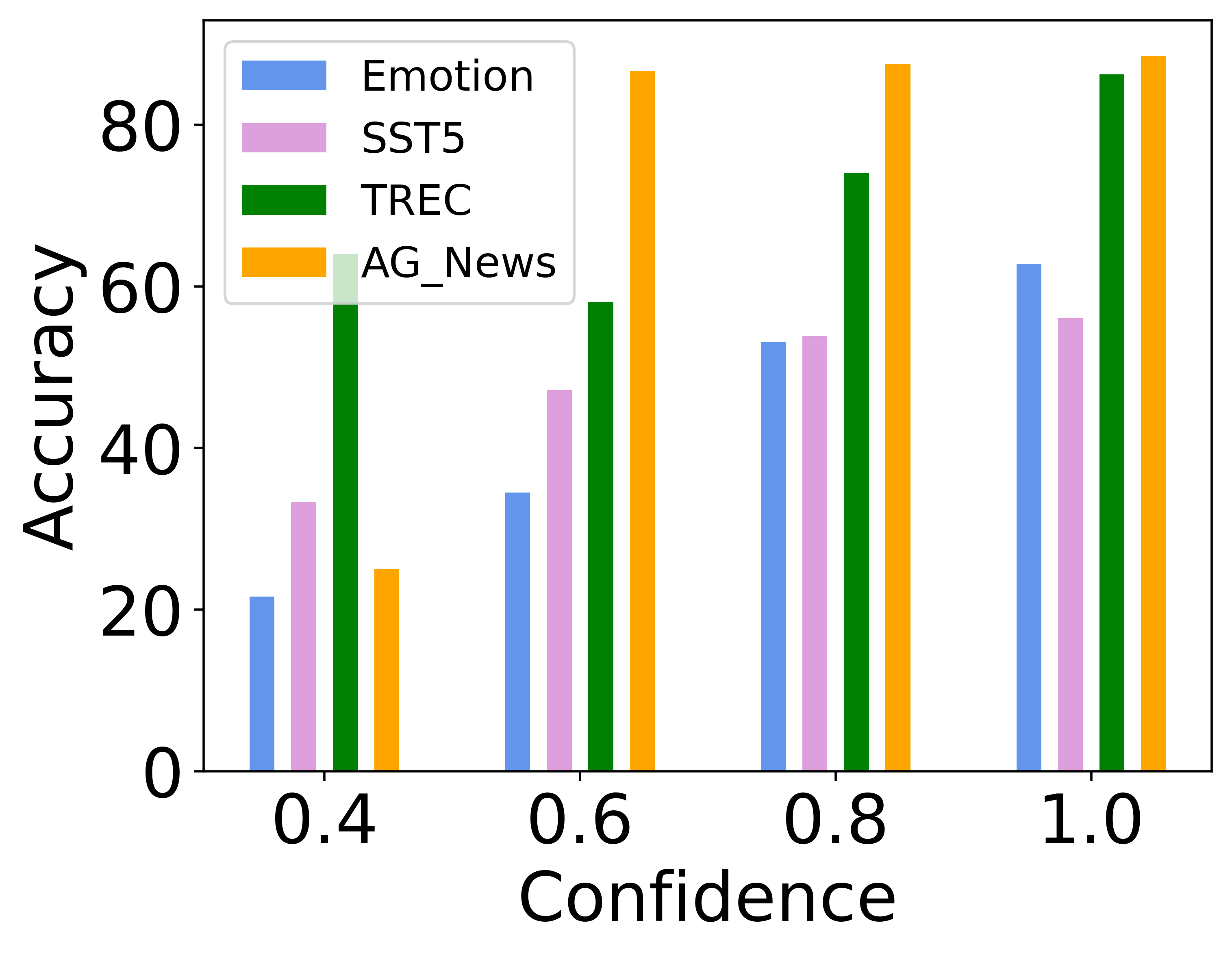}
    }
    \vspace{-3mm}
    \caption{Model Accuracy vs Voting consistency score using PaLM. We observe a similar positive correlation between voting consistency score and accuracy, thus making it a reliable confidence metric.} 
    \label{Voting cosistency VS model accuracy on PaLM}
    \vspace{-5mm}
\end{figure}

Figure~\ref{Voting cosistency VS model accuracy on PaLM} shows the analysis of adopting voting consistency as confidence score on PaLM.

\section{Prompt for Prompt-Ensemble}
\label{Prompt for Prompt-Ensemble}

Table~\ref{Prompts for Prompt-Ensemble} shows the manually written prompt we prepare for each dataset in the Prompt-Ensemble method.

\begin{table}[t!]
\centering
\begin{tabular}{l|l}
\hline
Dataset                  & Prompt                                                             \\ \hline
\multirow{5}{*}{SST5}    & Label the sentiment class of the sentence.                         \\ \cline{2-2} 
                         & What is the sentiment expressed in this message?                   \\ \cline{2-2} 
                         & What sentiment does this message express?                          \\ \cline{2-2} 
                         & How will you feel about the message in terms of its sentiment?     \\ \cline{2-2} 
                         & What sentiment does the writer express for the message?            \\ \hline
\multirow{5}{*}{TREC}    & Label the categories of the given question.                        \\ \cline{2-2} 
                         & What is the categories of the given question?                      \\ \cline{2-2} 
                         & What type of information does this question express?               \\ \cline{2-2} 
                         & How will you feel about the question about its category?           \\ \cline{2-2} 
                         & What type of information does the writer express for the question? \\ \hline
\multirow{5}{*}{Emotion} & Label the emotion class of the sentence.                           \\ \cline{2-2} 
                         & What is the emotion expressed in this message?                     \\ \cline{2-2} 
                         & What emotion does this message express?                            \\ \cline{2-2} 
                         & How will you feel about the message in terms of its emotion?       \\ \cline{2-2} 
                         & What emotion does the writer express for the message?              \\ \hline
\end{tabular}
\caption{Prompt we manually write for Prompt-Ensemble approach in each dataset.}
\label{Prompts for Prompt-Ensemble}
\end{table}


\end{document}